\documentclass{article}

\usepackage[preprint]{neurips_2025}

\usepackage[utf8]{inputenc} 
\usepackage[T1]{fontenc}    
\usepackage{hyperref}       
\usepackage{url}            
\usepackage{booktabs}       
\usepackage{amsfonts}       
\usepackage{nicefrac}       
\usepackage{microtype}      
\usepackage{xcolor}         
\usepackage{graphicx}
\usepackage{amsmath, amssymb, amsthm}
\usepackage{multirow}

\newtheorem{theorem}{Theorem}

\title{Understanding Transformer Encoder–Decoder Representations through Bernoulli Dropout}

\author{%
Xuanzhou Chen \\
  School of Electrical and Computer Engineering\\
  Georgia Institute of Technology\\
  Atlanta, GA 30332 \\
\texttt{xchen920@gatech.edu} \\
}

\begin{document}
\maketitle
\begin{abstract}
We study Transformer overparameterization through the lens of angular similarity in high-dimensional encoder–decoder embeddings. We apply Bernoulli dropout between the encoder and the decoder, varying the keep probability $p$ to identify a sparsity-dependent threshold above which the Top-1 prediction is preserved. Theoretically, we prove that, if the effective sparsity embeddings is sufficiently large, and thus decoder performance, remain stable under moderate coordinate dropout. Empirically, we implement the Bernoulli dropout by constructing a new Transformer model augmented with Binary Erasure Channel (BEC) and test its performance on an English–French translation task. Experimental results visualize the trends for validation accuracies and BLEU scores, both decline sharply at some threshold.
\end{abstract}

\section{Introduction}
Transformers are highly expressive, but this expressivity makes them prone to overfitting when the encoder–decoder interface carries rich, high-dimensional representations~\cite{https://doi.org/10.48550/arxiv.1706.03762, Varis_2021}. Prior work attributes this to a mismatch between model capacity and data scale, i.e, Transformers are overparameterized, where excessive parameters relative to training tokens induce memorization of spurious patterns rather than generalizable features~\cite{hoffmann2022trainingcomputeoptimallargelanguage}. Such overparameterization in Transfomer yields redundancy in information embedded in high-dimensional vectors.

To understand how much the representation of the embeddings is truly informative, we use Bernoulli dropout~\cite{srivastava2014dropout} to study the geometry by focusing on angular similarity structure of embeddings. Theoretically, we analyze whether the learned high-dimensional representations preserve their geometric structure under some sparsity condition. Specifically, we investigate if Top-1 prediction is preserved after applying Bernoulli dropout with different probabilities on the representation embeddings between the encoder and decoder. To evaluate the theoretical results empirically, we implement Bernoulli dropout via a Binary Erasure Channel (BEC) and compare a BEC-augmented Transformer with the base model on a neural machine translation task.  By probing the erasure probability, we control the sparsification in high-dimensional embedding vectors. The experimental results show consistency with our theoretical results.
\section{Related work}
In this section, we review theoretical results relevant to our study, including overparameterization, high-dimensional and representation. These works provide the mathematical basis for our use of Bernoulli dropout as a probe of geometric structure preservation in Transformer encoder--decoder embeddings. These following three lines of work motivate a central question: when erasing most (e.g., 90–95\%) of embedding coordinates, the primary concern is not computational cost but signal preservation. Does such aggressive sparsification compromise some geometry representation that is essential for the decoder’s accurate token prediction?

\paragraph{Overparameterization} Overparameterization in modern deep networks has been extensively studied in statistical learning theory. Classical bounds based on VC dimension or Rademacher complexity predict poor generalization when the number of parameters far exceeds the number of training samples. However, recent work on \emph{benign overfitting} (~\cite{Bartlett_2020,tsigler2022benignoverfittingridgeregression,bartlett2021deeplearningstatisticalviewpoint}) shows that, in certain structured settings, overparameterized models can interpolate the training data while still achieving low test error. 

\paragraph{High-dimensional representation}In high-dimensional settings, geometric concentration phenomena and stochastic separation theorems (~\cite{Gorban_2018}) reveal that randomly distributed points remain almost-orthogonal and linearly separable with high probability. When the encoder decoder architecture in a Transformer carries a rich, high-dimensional representation, these result may imply that even sparse or partially corrupted features may preserve discriminative structure, provided the corruption is sufficiently mild. 

\paragraph{Dropout inspired from information theory and learning theory} In this work, we focus specifically on Bernoulli-mask-based dropout between the encoder and the decoder by exploring a hybrid approach that draws inspiration from joint source–channel coding~\cite{DBLP:journals/corr/abs-1802-06832} in information theory. Instead of transmitting dense representations from encoder to decoder, we propose to insert a learnable channel layer that selectively filters and compresses the encoder output before decoding. Our goal is to retain essential semantic information through a sparse, lossy representation—thus reducing computational load—while preserving the decoder's ability to recover high-quality outputs. From a classical learning-theory view, dropout reduces the effective capacity of the hypothesis class—lowering data-dependent measures such as Rademacher complexity, and thus improves generalization. From a PAC-Bayes perspective, multiplicative Bernoulli noise implements a data-dependent posterior over predictors that tightens stochastic generalization bounds compared with deterministic decoders. From an Information Bottleneck (IB) perspective, dropout limits the mutual information transmitted across the interface, forcing the model to retain only task-relevant content while discarding nuisance variability.
\section{Methodology}

\begin{figure}[h]
    \centering
    \begin{minipage}[t]{0.45\textwidth}
        \centering
        \includegraphics[width=0.85\linewidth]{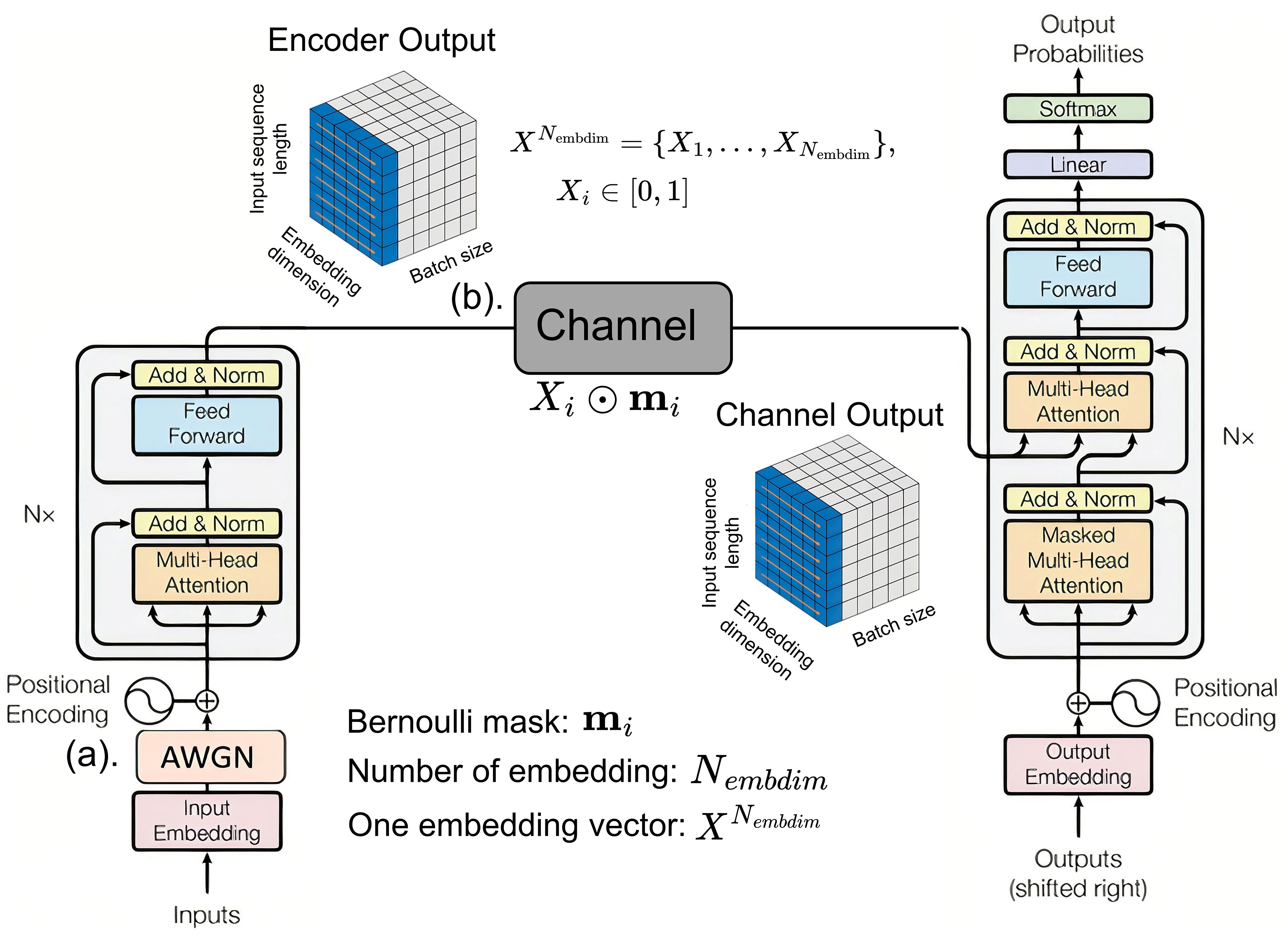}
        \caption{\small BEC-augmented Transformer architecture: The BEC between the encoder and decoder applies the Bernoulli mask $\mathbf{m}_i$. During training, Additive White Gaussian Noise (AWGN) is applied to perturb the input embeddings.}
        \label{fig:model_arch}
    \end{minipage}
    \hfill
    \begin{minipage}[t]{0.48\textwidth}
        \centering
        \includegraphics[width=0.85\linewidth]{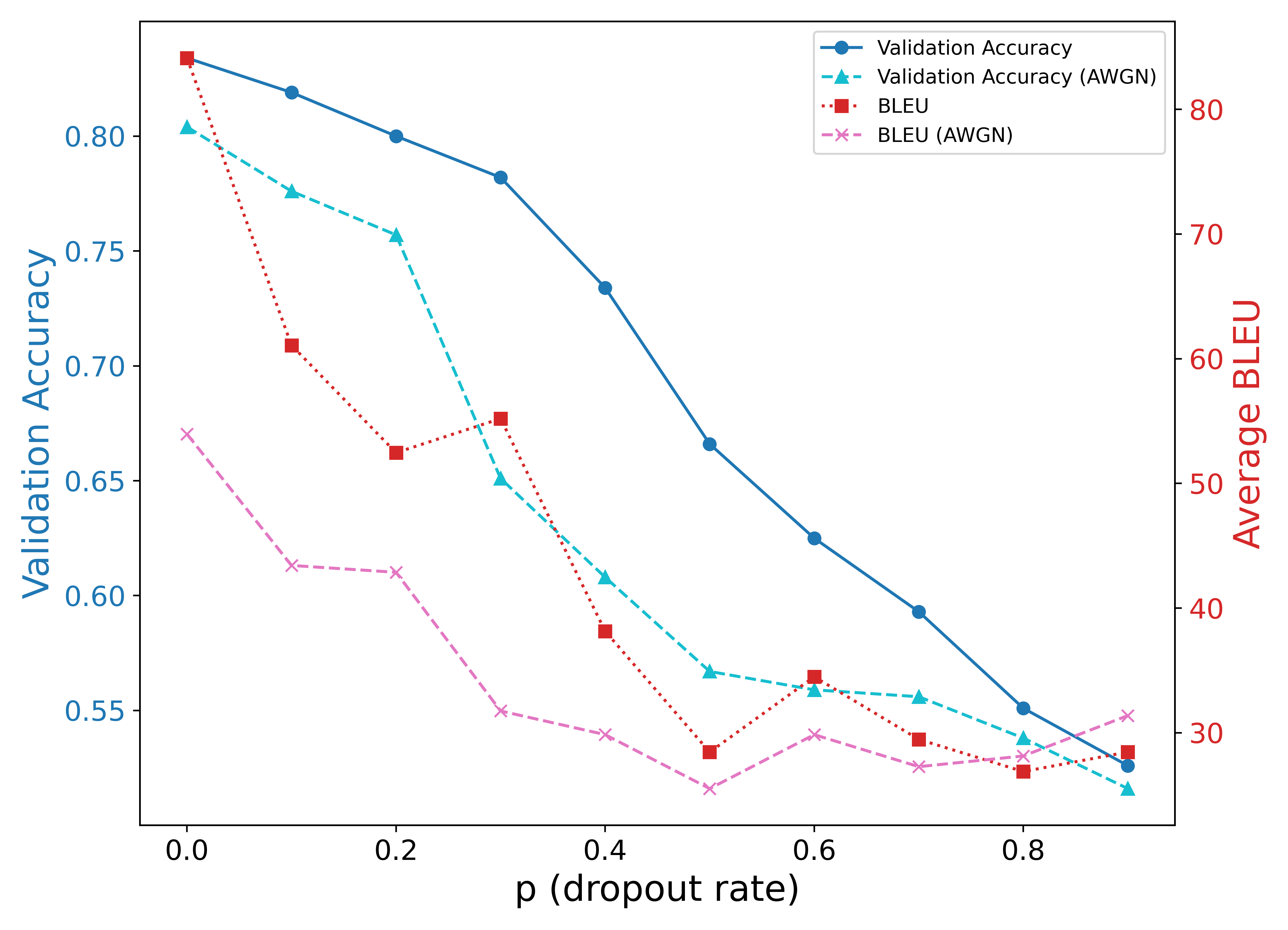}
        \caption{Validation accuracy and Average BLEU (per sentence) across different Bernoulli dropout probabilities ($p$) under both noise-free and AWGN training settings.}
        \label{fig:acc_plot}
    \end{minipage}
\end{figure}

Based on the encoder–decoder architecture of the Transformer, we aim to randomly drop entries in each embedding vector between the encoder and decoder (see details in Figure~\ref{fig:model_arch}). Mathematically, let $X_1, \ldots, X_M \in \mathbb{R}^{d}$ as $d$-dimensional embedding vectors in the encoder output, we apply a Bernoulli mask $\mathbf{m}_i \in \{0, 1\}^d$ with probability $p$ on each embedding vector in entrywise manner 
\[
[\mathbf{m}_i]_j = \begin{cases}
    1 & \text{  if  } |[X_i]_j| \geq p,\\
    0 & \text{otherwise.}
\end{cases}  \quad \text{    for } j = 1,\ldots,d.
\] We denote dropout output as $\tilde{X}_i$, where \(
\tilde{X}_i = X_i \odot \mathbf{m}_i.
\) 
Inspired by the channel modeling~\cite{DBLP:journals/corr/abs-1908-05731} in neural machine translation. We implement the dropout via Binary Erasure Channel (BEC). Particularly, for each embedding vector which is denoted as $X^{N_{\text{embdim}}}=\{ X_{1}, ..., X_{N_{\text{embdim}}}\}$, $X_{i} \in [0,1]$, we define BEC with probability $p$ as the coordinate-wise function $g$ where

\begin{equation}\label{eq:def_binary_mask}
g([X_i]_j) = 
\begin{cases} 
[X_i]_j, & [X_i]_j \geq p \\
0, & [X_i]_j < p \\
\end{cases}.
\end{equation}

The channel is inserted between the encoder and decoder, with the number of parallel channels equal to the product of the input sequence length and batch size.

\paragraph{Add AWGN noise} We then choose to apply additive white Gaussian noise (AWGN) after the input embedding step. AWGN is a widely used and analytically tractable noise model whose statistical properties are well understood, making it an ideal choice for controlled perturbation experiments. In our context, applying AWGN to the embeddings allows us to simulate continuous-valued channel noise with a specified variance, perturbing all coordinates in an unbiased and isotropic manner. This property is important because it preserves the overall symmetry of the embedding space while introducing realistic distortions, enabling us to isolate the effect of noise magnitude on the geometric structure of the representations. Moreover, AWGN serves as a baseline perturbation model in communication systems and signal processing, allowing us to connect our analysis with established theoretical results on robustness and capacity under Gaussian noise. By introducing AWGN at the embedding stage, we can systematically probe how much geometric information and decoder accuracy can be preserved when the representation is subject to uniform stochastic interference.

If $p$ is close to 1, then this process retains entries with magnitude approximately at $p$ and zeros out the rest, resulting in a sparse representation. After dropout, the embedding vectors are normalized and fed into the Transformer decoder. 

\paragraph{Theoretical results} To study the extent to which dropout affects the decoder’s decision,
we pose the following research question:
\textit{Given a query vector $q$ and output embeddings $\{v_j\}_{j=1}^M$ with a positive pre-dropout margin $\gamma$,
does the top-1 prediction $\arg\max_j \langle q, v_j\rangle$ remain unchanged with high probability
when the post-dropout keep size $s=p\,d$ is moderately large (e.g., $s\gg \log M$)?}
To address this question, we present the following result.

\begin{theorem}[Top-1 prediction preserved under coordinate dropout, proved in Appendix~\ref{app:pf_thm}]
\label{thm:top1-preserve}
Let $q\in\mathbb{R}^d$ be the decoder state just before the output projection and
$\{v_j\}_{j=1}^M\subset\mathbb{R}^d$ be unit output embeddings. Assume $\|q\|_2=1$ and define
$s_j=\langle q,v_j\rangle$, $j^\star=\arg\max_{j\in[M]} s_j$, and the margin
$\gamma:= s_{j^\star}-\max_{j\neq j^\star} s_j>0$. Draw a Bernoulli mask
$M=\mathrm{diag}(m_1,\ldots,m_d)$ with $m_k\stackrel{\text{i.i.d.}}{\sim}\mathrm{Ber}(p)$,
set $\tilde q = M q$ and $\hat q=\tilde q/\|\tilde q\|$ (if $\tilde q\neq 0$).
Let the effective sparsity be $s_{\mathrm{eff}}(q):=\|q\|_2^4/\|q\|_4^4\in[1,d]$. Then there exists a universal constant $C>0$ such that, for any $\delta\in(0,1)$,
with probability at least $1-\delta$,
\[
\max_{j\in[M]}\,\big|\langle \hat q,v_j\rangle-\langle q,v_j\rangle\big|
\;\le\; C\,\sqrt{\frac{\log(M/\delta)}{p\,s_{\mathrm{eff}}(q)}}.
\]
Consequently, if
\(
\gamma>2C \sqrt{\frac{\log(M/\delta)}{ps_{\mathrm{eff}}(q)}},
\)
then the post-dropout prediction is unchanged:
\(
\arg\max_{j\in[M]} \langle \hat q, v_j\rangle=j^\star
\) with probability at least $1-\delta$.
\end{theorem}
Theorem~\ref{thm:top1-preserve} states that if the pre-dropout decoder margin $\gamma$ is positive and 
the effective sparsity $s_{\min}$ is sufficiently large, then Bernoulli dropout with a shared mask preserves the top-1 prediction with high probability. 
In particular, the probability of a prediction change decays exponentially in $s_{\min}\,\gamma^2$, up to a $\log(M/\delta)$ factor.

\section{Experiments}

\paragraph{Data pre-processing} We use this Eng-Fre translation corpus from pytorch tutorial as our dataset. The dataset is a collection of thousands of English-to-French translation pairs. Each sentence pair consists of one English sentence and its corresponding target sentence in French. The data preparation proceeds as follows: 1) split the text files into lines and then pair the lines; 2) normalize the text and filter the pairs by length and content; 3) construct a vocabulary list from the resulting sentence pairs. Sentences from the corpus are padded and truncated up to 50 tokens, one-hot encoded, and trained with teacher forcing (0.5).

\paragraph{Experiment setup}Noise is added as perturbation on the input via an AWGN layer. The dropout probability $p$ for BEC is varied to measure its impact on test accuracies and BLEU scores, assessing whether Top-1 prediction is preserved after dropout. Figure~\ref{fig:acc_plot} visualizes the trends of validation accuracy and BLEU under both noise-free and AWGN setting.
In our training process, we set the maximal sentence length to 50 tokens with padding for short sentences and truncation for long sentences. One-hot coding is applied and the total word count is limited to be less than thousands. We also use "teacher forcing" to faster the convergence of the model with the teacher forcing ratio at 0.5.
\begin{figure}[!t]
    \centering
    \includegraphics[width=9cm]{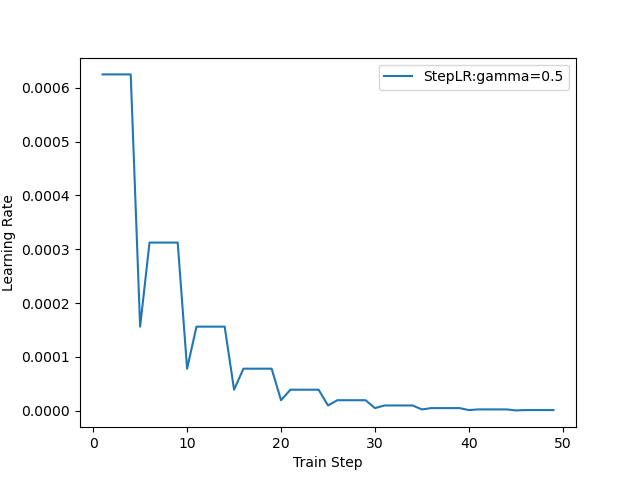}
    \caption{\parbox{0.7\linewidth}{\centering
        Scheduled learning rate in Adam for noise-free and AWGN training.
    }}
    \label{fig:schedule_lr}
\end{figure}
In both Noise-free (no AWGN) and AWGN experiments, we set the batch size to be 64 and the epoch to be 80. For all training experiments, we adopt Adam as optimizer (betas=($0.9, 0.98$), $eps=1e-9$) with scheduled learning rate shown in Figure~\ref{fig:schedule_lr}. All experiments are run on a Windows 11 machine with 32.00 GB (3200 MHz) RAM, NVIDIA GeForce RTX 3060 Laptop GPU (4095MB) and 11th Gen Intel(R) Core(TM) i7-11800H @ 2.30GHz CPU. We compare original Transformer and BEC-augmented Transformers with different dropout probability $p$. 

\subsection{Numerical Results}
We first present our results on both training time and validation accuracies and BLEU scores while sweeping over different values of dropout probability $p$.

\begin{figure}[htp]
    \centering
    \includegraphics[width=14cm]{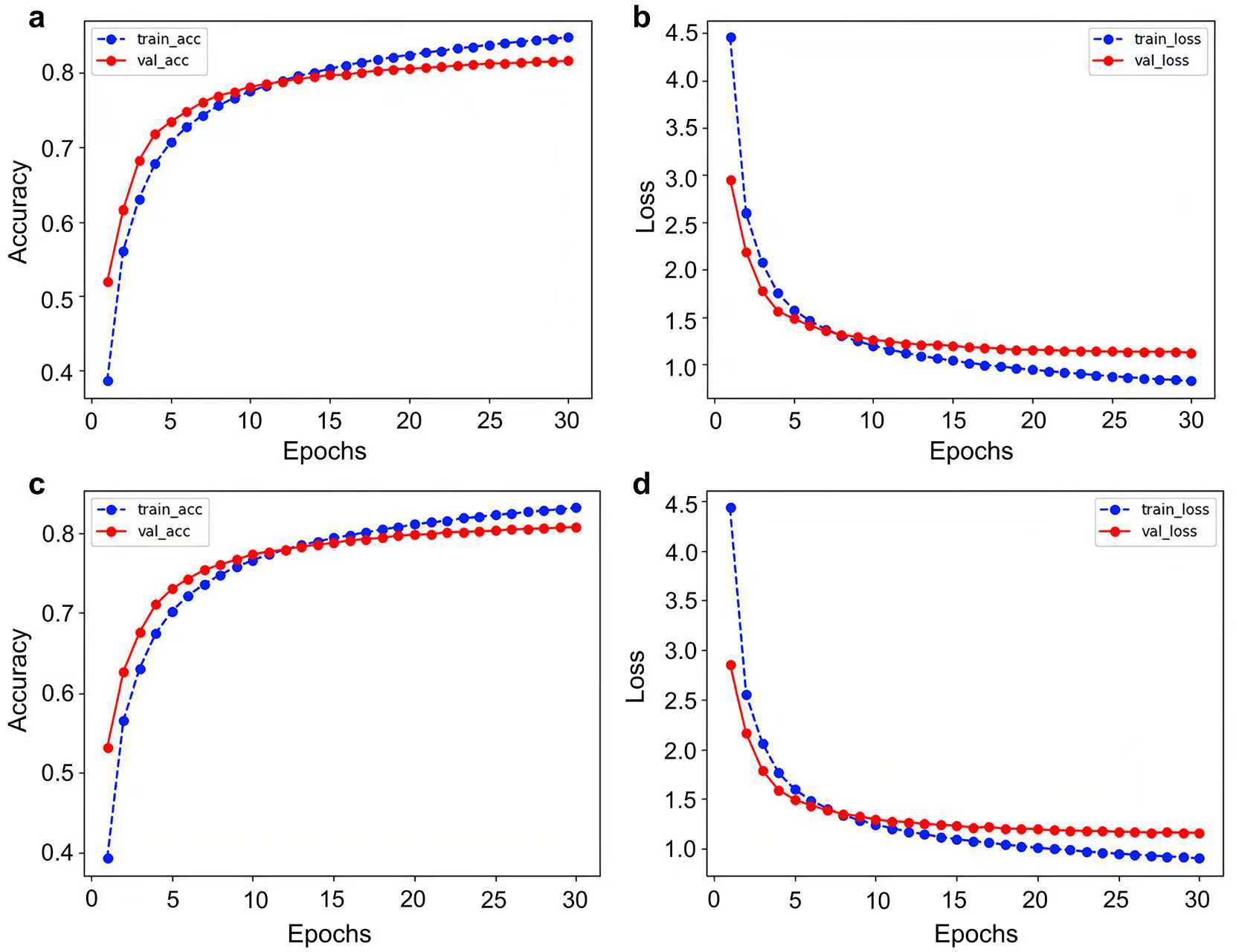}
    \caption{(a) Training and testing accuracies for the BEC-augmented Transformer with dropout probability $0.2$ in the noise-free experiment. (b) Training and testing losses for the BEC-augmented Transformer with dropout probability $0.2$ in the noise-free experiment. (c) Training and testing accuracies for the BEC-augmented Transformer with dropout probability $0.5$ in the noise-free experiment. (d)Training and testing losses for the BEC-augmented Transformer with dropout probability $0.5$ in the noise-free experiment.}
    \label{fig:acc_loss_plot}
\end{figure}

\paragraph{Training time and convergence} Table \ref{tab:time-table} shows the computational time required for the baseline Transformer and the BEC-augmented models under the AWGN setting. Both BEC(0.2) and BEC(0.5) reduce training time compared to the baseline, achieving speedups of approximately 5.7\% and 7.7\%, respectively. This reduction is likely due to the sparsification introduced by the binary erasure channel, which decreases the number of active computations during forward and backward passes. Notably, despite this efficiency gain, the accuracy drop observed remains relatively small, suggesting that BEC can provide computational benefits without severely compromising performance. We examine the convergence behavior during the first 30 epochs and find that the training progresses at nearly the same rate for 
$p=0.2$ and $p=0.5$. This suggests that, within this early phase, increasing the dropout probability does not hinder the model’s ability to reduce the training loss.
\begin{table}[h]
  \centering
  \begin{tabular}{lllll}
    \toprule
    Model     & Noise     & Batch Size & Epochs & Time(s) \\
    \midrule
    Transformer & AWGN & \quad 128 & 80 &  9650\\
    Transformer + BEC(0.2)  & AWGN &  \quad 128 & 80 & 9101\\
    Transformer + BEC(0.5)  & AWGN &  \quad 128 & 80 & 8906\\
    \bottomrule
 \end{tabular}
 \vspace{0.3em}
    \caption{Training Time for baseline Transformer and BEC-augmented Transformers under AWGN training.}
    \label{tab:time-table}
     
\end{table}

\begin{table}[htp]
\centering
\begin{tabular}{llccc}
\hline
\textbf{Model} & \textbf{Dropout $p$} & \textbf{AWGN} & \textbf{Validation Accuracy} & \textbf{BLEU} \\
\hline
\multirow{2}{*}{Transformer} & \multirow{2}{*}{0.0} 
& No  & $0.834$ & $84.12$ \\
& & Yes & $0.804(\downarrow 3.6\%)$ & $53.96(\downarrow 35.9\%)$ \\
\hline
\multirow{2}{*}{Transformer + BEC} & \multirow{2}{*}{0.1} 
& No  & $0.819(\downarrow 1.8\%)$ & $61.08(\downarrow 27.4\%)$ \\
& & Yes & $0.776(\downarrow 7.0\%)$ & $43.42(\downarrow 48.4\%)$ \\
\hline
\multirow{2}{*}{Transformer + BEC} & \multirow{2}{*}{0.2} 
& No  & $0.800(\downarrow 4.1\%)$ & $52.46(\downarrow 37.6\%)$ \\
& & Yes & $0.757(\downarrow 9.2\%)$ & $42.87(\downarrow 49.0\%)$ \\
\hline
\multirow{2}{*}{Transformer + BEC} & \multirow{2}{*}{0.3} 
& No  & $0.782(\downarrow 6.2\%)$ & $55.20(\downarrow 34.4\%)$ \\
& & Yes & $0.651(\downarrow 21.9\%)$ & $31.75(\downarrow 62.3\%)$ \\
\hline
\multirow{2}{*}{Transformer + BEC} & \multirow{2}{*}{0.4} 
& No  & $0.734(\downarrow 12.0\%)$ & $38.15(\downarrow 54.6\%)$ \\
& & Yes & $0.608(\downarrow 27.1\%)$ & $29.85(\downarrow 64.5\%)$ \\
\hline
\multirow{2}{*}{Transformer + BEC} & \multirow{2}{*}{0.5} 
& No  & $0.666(\downarrow 20.1\%)$ & $28.45(\downarrow 66.2\%)$ \\
& & Yes & $0.567(\downarrow 32.0\%)$ & $25.52(\downarrow 69.7\%)$ \\
\hline
\multirow{2}{*}{Transformer + BEC} & \multirow{2}{*}{0.6} 
& No  & $0.625(\downarrow 25.1\%)$ & $34.49(\downarrow 59.0\%)$ \\
& & Yes & $0.559(\downarrow 33.0\%)$ & $29.85(\downarrow 64.5\%)$ \\
\hline
\multirow{2}{*}{Transformer + BEC} & \multirow{2}{*}{0.7} 
& No  & $0.593(\downarrow 28.9\%)$ & $29.47(\downarrow 65.0\%)$ \\
& & Yes & $0.556(\downarrow 33.3\%)$ & $27.28(\downarrow 67.6\%)$ \\
\hline
\multirow{2}{*}{Transformer + BEC} & \multirow{2}{*}{0.8} 
& No  & $0.551(\downarrow 33.9\%)$ & $26.89(\downarrow 68.0\%)$ \\
& & Yes & $0.538(\downarrow 35.5\%)$ & $28.14(\downarrow 66.5\%)$ \\
\hline
\multirow{2}{*}{Transformer + BEC} & \multirow{2}{*}{0.9} 
& No  & $0.526(\downarrow 36.9\%)$ & $28.46(\downarrow 66.2\%)$ \\
& & Yes & $0.516(\downarrow 38.1\%)$ & $31.38(\downarrow 62.7\%)$ \\
\hline
\end{tabular}
\vspace{0.3em}
\caption{Validation accuracies and BLEU scores for the baseline Transformer model and BEC-augmented Transformer across dropout probabilities ranging from 0.0 to 0.9, evaluated under both noise-free and AWGN training conditions. Percentage changes in validation accuracy (related to the baseline, $p=0$) is also reported.}
\label{tab:dropout_experiments}
\end{table}
\paragraph{Test accuracy and BLEU evaluation} Table~\ref{tab:dropout_experiments} reports the test accuracy and BLEU scores obtained by varying the Bernoulli keep probability. From the results, we observe a clear downward trend in both metrics as $p$ increases (i.e., as dropout strength increases) for both the noise-free and AWGN conditions. Furthermore, performance in the AWGN setting is consistently lower than in the noise-free setting.

\subsection{Semantic Results} 
We then assess semantic meaning preservation by examining all translation sentence pairs. Since loss and accuracy alone are insufficient to fully assess the preservation of semantic meaning in the translation results, we additionally present sample translations and conduct a human evaluation. In this section, we evaluate by  comparing semantic meaning in translation results for both short and complex sentences under different parameters settings, denoted in the format of pred(AWGN mean, BSC probability). pred(original) means the prediction from originally trained transformer model.

\textbf{Sample sentences (short)}

< Sample 1 > \\
input: je pars en vacances pour quelques jours . \\
target: i m taking a couple of days off . \\
pred(original):      <start>i m taking a couple of days off .\\
pred(0.1, 0.8): <start> i m taking a couple of days off . \\
pred(0.0, 0.5): <start> i m taking a vacation . \\

< Sample 2 > \\
input: je ne me panique pas . \\
target: i m not panicking . \\
pred(original):      <start> i m not panicking .\\
pred(0.1, 0.8): <start> i m not panicking .\\
pred(0.0, 0.5): <start> i m not panicking .\\

< Sample 3 > \\
input: je recherche un assistant .\\
target: i am looking for an assistant .\\
pred(original):      <start> i am looking for an assistant .\\
pred(0.1, 0.8): <start> i am looking for an assistant .\\
pred(0.0, 0.5): <start> i m looking for a sweater .\\

< Sample 4 > \\
input: je suis loin de chez moi .\\
target: i m a long way from home .\\
pred(original):      <start> i m a long way from home .\\
pred(0.1, 0.8): <start> i m a long way from home .\\
pred(0.0, 0.5): <start> i m tired of all this nagging .\\

< Sample 5 > \\
input: vous etes en retard .\\
target: you re very late .\\
pred(original):      <start> you are very late .\\
pred(0.1, 0.8): <start> you are late .\\
pred(0.0, 0.5): <start> you re really absent minded .\\

< Sample 6 > \\
input: j ai soif .\\
target: i am thirsty .\\
pred(original):      <start> i am thirsty .\\
pred(0.1, 0.8): <start> i m thirsty .\\
pred(0.0, 0.5): <start> i m done .\\

< Sample 7 > \\
input: je suis fou de vous .\\
target: i m crazy about you .\\
pred(original):      <start> i m crazy about you .\\
pred(0.1, 0.8): <start> i m crazy about you .\\
pred(0.0, 0.5): <start> i m lucky to have you as a friend .\\

< Sample 8 > \\
input: vous etes vilain .\\
target: you are naughty .\\
pred(original):      <start> you re bad .\\
pred(0.1, 0.8): <start> you re bad .\\
pred(0.0, 0.5): <start> you re grumpy .\\

< Sample 9 > \\
input: il est vieux et laid .\\
target: he s old and ugly .\\
pred(original):      <start> he s old and ugly .\\
pred(0.1, 0.8): <start> he s old and ugly .\\
pred(0.0, 0.5): <start> he s old .\\

< Sample 10 > \\
input: je suis terrifiee .\\
target: i m terrified .\\
pred(original):      <start> i m hit .\\
pred(0.1, 0.8): <start> i m a desperate .\\
pred(0.0, 0.5): <start> i m wet .\\

In above prediction results, pred(original) achieved the testing accuracy at $0.822$ after 100 training epochs, while pred($0.1, 0.1, 0.8$), pred($0.1, 0.0, 0.5$) achieved the test accuracies at $0.821$ and $0.723$, respectively.

We can see the trade-off between the semantic meaning preservation and the testing accuracy. For example, from the testing result of pred(0.1, 0.0, 0.5), we observe that when the model is trained with stronger noise, it mostly preserves the semantic meaning but intends to translate with substitution of words in the target sentence (see sample 8), or even substitution of phrases and sentences (sample 1 \& 7). However, strong noise may corrupt the training sample and cause the model to learn from wrong or partial information (see sample 3).

To further examine the performance, we compare the model with the same set of parameters with using sentences with more complex grammar and semantic meaning. \\

\textbf{Sample sentences (complex)}

< Sample 1 > \\
input: Vous allez le briser si vous ne faites pas attention . \\
target: You ll break it if you re not careful . \\
pred(original):      <start> you re not going to do it  .\\
pred(0.1, 0.8): <start>  you re not going to keep it . \\

< Sample 2 > \\
input: Je n aurais jamais pensé qu il y ait là un tel endroit tranquille dans cette ville bruyante . \\
target: I never dreamed of there being such a quiet place in this noisy city . \\
pred(original):      <start> i m not thinking it was in a new ideas .\\
pred(0.1, 0.8): <start> i m not thinking about this until now.\\

< Sample 3 > \\
input: Votre voix me semble très familière .\\
target: Your voice sounds very familiar to me .\\
pred(original):      <start> i m really stranger to your voice .\\
pred(0.1, 0.8): <start> i m very happy to hear your voice .\\

< Sample 4 > \\
input: Presque tout le monde est déjà rentré à la maison .\\
target: Almost everybody has already gone home .\\
pred(original):      <start> i m almost home all the time .\\
pred(0.1, 0.8): <start> we re almost to the same house .\\

\begin{figure}[htp]
    \centering
    \includegraphics[width=14cm]{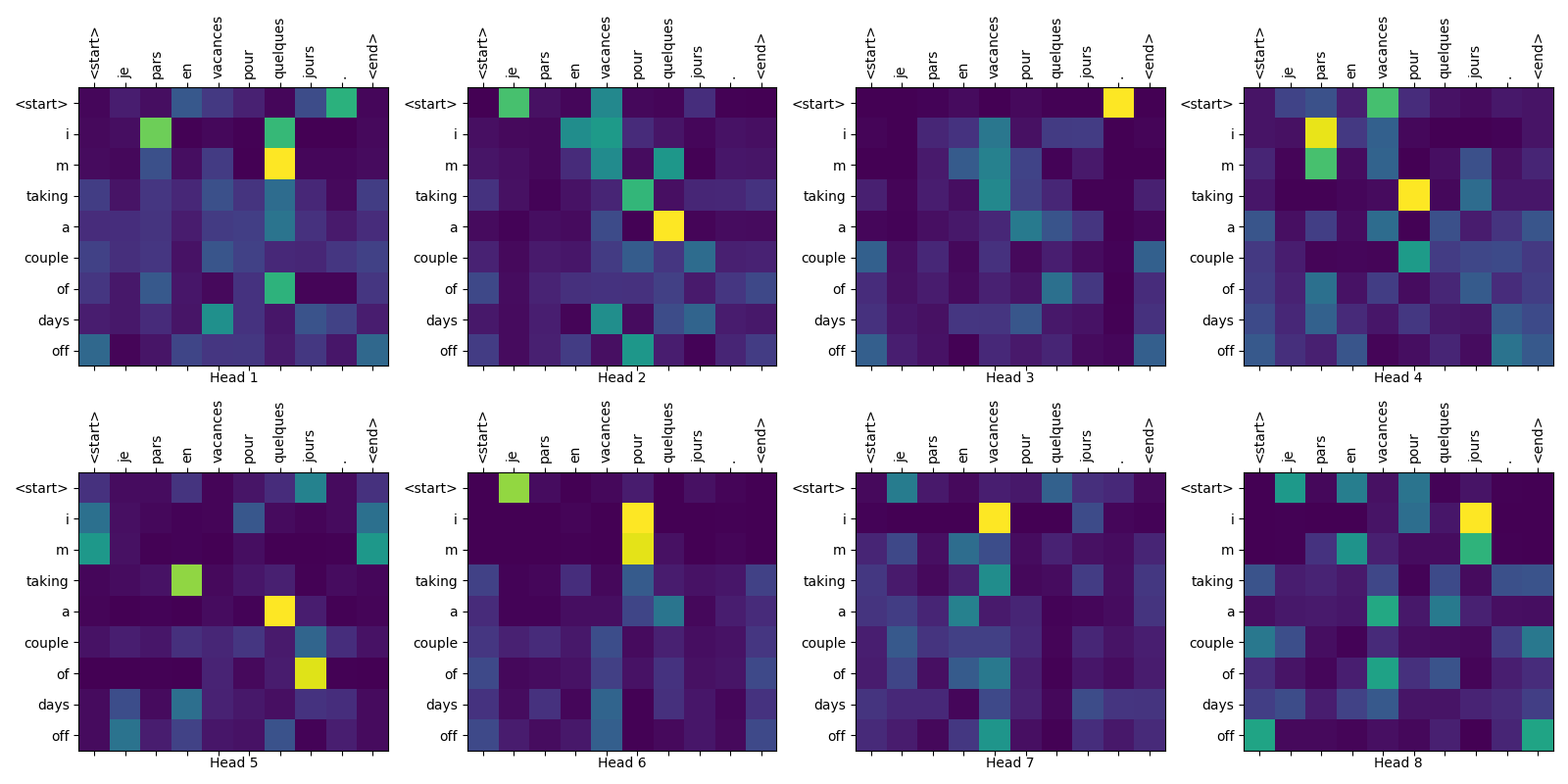}
    \caption{Attention visualization heatmap for BEC-augmented Transformer with dropout probability at $0.5$.}
    \label{fig:heatmap}
\end{figure}

We observe that although the originally trained Transformer model outperforms on preserving grammars of the target language, it does not preserve the semantic meaning of the sentence. It could negate the meaning or refer to unrelated information. In contrast, although Transformer model incorporating a BEC and trained under AWGN does not preserve grammar of the target language, it preserves the semantic meaning in a subtle way. Finally, the heat map in Figure~\ref{fig:heatmap} illustrates that, with multi-head attention, the BEC-augmented model can still focus on relevant locations within sentence pairs, even when those locations are nonconsecutive or far apart.

\section{Conclusion}
We have presented a theoretical and empirical analysis of Transformer representations through the lens of Bernoulli dropout. Our experiments reveal that encoder–decoder embeddings exhibit a high degree of redundancy, maintaining the angular similarity required for correct predictions even under high sparsity conditions. The identification of a sharp breakdown threshold supports our theoretical bounds on effective sparsity. This work not only characterizes the geometry of Transformer overparameterization but also highlights the potential for developing more parameter-efficient models by leveraging the inherent robustness of high-dimensional semantic spaces.
\small
\bibliographystyle{unsrt}
\bibliography{ref.bib}


\newpage
\appendix

\section{Technical Appendices and Supplementary Material}

\subsection{Appendix B: Proof of Theorem~\ref{thm:top1-preserve}}\label{app:pf_thm}
Write the post-dropout (pre-normalization) score and norm as
\[
S_j \;:=\; \langle M q, v_j\rangle \;=\; \sum_{k=1}^d m_k q_k v_{j,k},
\qquad
R^2 \;:=\; \|M q\|_2^2 \;=\; \sum_{k=1}^d m_k q_k^2,
\]
where $m_k \stackrel{\text{i.i.d.}}{\sim}\mathrm{Ber}(p)$ and $\|q\|_2=\|v_j\|_2=1$.
Let $s_{\mathrm{eff}}(q):=\|q\|_2^4/\|q\|_4^4 \in [1,d]$.

\medskip
\noindent\emph{Step 1: Concentration for the numerators $\{S_j\}$.}
For fixed $j$, define mean-zero summands $Y_{k}^{(j)}:=(m_k-p) q_k v_{j,k}$, so that
$S_j - p\langle q,v_j\rangle = \sum_k Y_k^{(j)}$.
Each $Y_k^{(j)}$ is independent, bounded by $|q_k v_{j,k}|$, and has variance
$\text{Var}(Y_k^{(j)}) = p(1-p) q_k^2 v_{j,k}^2 \le p\,q_k^2 v_{j,k}^2$.
By Bernstein’s inequality, for all $t>0$,
\[
\Pr\!\left(\big|S_j - p\langle q,v_j\rangle\big|\ge t\right)
\;\le\;
2\exp\!\left(
-\frac{c\,t^2}{p\sum_k q_k^2 v_{j,k}^2 + t\max_k |q_k v_{j,k}|}
\right).
\]
Using Cauchy–Schwarz, $\sum_k q_k^2 v_{j,k}^2 \le \|q\|_4^2 \|v_j\|_4^2 \le \|q\|_4^2$.
Taking a union bound over $j\in[M]$ and choosing
$t = C\sqrt{p\,\|q\|_4^2 \log(M/\delta)}$,
we obtain with probability at least $1-\delta/2$ that, simultaneously for all $j$,
\begin{equation}
\label{eq:num}
\big|S_j - p\langle q,v_j\rangle\big|
\;\le\; C\sqrt{p\,\|q\|_4^2 \log(M/\delta)}.
\end{equation}
(When $p\,s_{\mathrm{eff}}(q)\gtrsim \log(M/\delta)$ the variance term dominates and the linear term in Bernstein can be absorbed into the same rate.)

\medskip
\noindent\emph{Step 2: Concentration for the denominator $R$.}
Set $Z_k := m_k q_k^2$ (independent, bounded by $q_k^2$) with mean
$\mathbb{E} Z_k = p q_k^2$ and variance $\text{Var}(Z_k) \le p q_k^4$.
By Bernstein’s inequality applied to $\sum_k (Z_k - \mathbb{E} Z_k)$, for any $u>0$,
\[
\Pr\!\left(\big|R^2 - p\big|\ge u\right)
\;\le\;
2\mathbb{E}xp\!\left(
-\frac{c\,u^2}{p\sum_k q_k^4 + u\max_k q_k^2}
\right)
\;=\;
2\exp\!\left(
-\frac{c\,u^2}{p\,\|q\|_4^4 + u\,\|q\|_\infty^2}
\right).
\]
Choose $u = C\sqrt{p\,\|q\|_4^4 \log(2/\delta)}$; then with probability at least $1-\delta/2$,
\[
\big|R^2 - p\big| \;\le\; C\sqrt{p\,\|q\|_4^4 \log(2/\delta)}.
\]
A standard translation from $R^2$ to $R$ (mean-value inequality) yields
\begin{equation}
\label{eq:den}
\Big|R - \sqrt{p}\,\Big| \;\le\; C\,\sqrt{\frac{\|q\|_4^4}{p}}\,\sqrt{\log(2/\delta)}
\;=\; C\,\sqrt{\frac{\log(2/\delta)}{p\,s_{\mathrm{eff}}(q)}}.
\end{equation}
Consequently, for $p\,s_{\mathrm{eff}}(q)\gtrsim \log(1/\delta)$ we have the lower bound
\begin{equation}
\label{eq:Rlower}
R \;\ge\; \sqrt{p}\,\Big(1 - C\sqrt{\tfrac{\log(2/\delta)}{p\,s_{\mathrm{eff}}(q)}}\Big).
\end{equation}

\medskip
\noindent\emph{Step 3: Uniform score deviation after normalization.}
Post-dropout (normalized) scores are $s'_j=\langle \hat q, v_j\rangle = S_j/R$.
Decompose
\[
s'_j - \langle q,v_j\rangle
= \frac{S_j - p\langle q,v_j\rangle}{R}
+ \langle q,v_j\rangle\left(\frac{p}{R}-1\right).
\]
Using \eqref{eq:num} and \eqref{eq:Rlower},
\[
\left|\frac{S_j - p\langle q,v_j\rangle}{R}\right|
\;\le\;
\frac{C\sqrt{p\,\|q\|_4^2 \log(M/\delta)}}{\sqrt{p}\,\big(1 - C\sqrt{\tfrac{\log(2/\delta)}{p\,s_{\mathrm{eff}}(q)}}\big)}
\;\le\;
C'\,\sqrt{\frac{\log(M/\delta)}{s_{\mathrm{eff}}(q)}},
\]
and from \eqref{eq:den},
\[
\left|\frac{p}{R}-1\right|
= \frac{|p-R|}{R}
\;\le\;
\frac{C\,\sqrt{p\,\|q\|_4^4 \log(2/\delta)}}{\sqrt{p}\,\big(1 - C\sqrt{\tfrac{\log(2/\delta)}{p\,s_{\mathrm{eff}}(q)}}\big)}
\;\le\;
C'\,\sqrt{\frac{\log(1/\delta)}{p\,s_{\mathrm{eff}}(q)}}.
\]
Since $|\langle q,v_j\rangle|\le 1$, combining the two bounds and union-bounding over $j\in[M]$ gives, with probability at least $1-\delta$,
\begin{equation}
\label{eq:uniform-score}
\max_{j\in[M]}\big|s'_j - s_j\big|
\;\le\;
C\,\sqrt{\frac{\log(M/\delta)}{p\,s_{\mathrm{eff}}(q)}}.
\end{equation}

\medskip
\noindent\emph{Step 4: Margin preservation and top-1 invariance.}
Let $j^\star=\arg\max_j s_j$ and $\gamma = s_{j^\star}-\max_{j\ne j^\star} s_j>0$.
From \eqref{eq:uniform-score},
\[
s'_{j^\star} - \max_{j\ne j^\star} s'_j
\;\ge\;
\big(s_{j^\star}-\varepsilon\big) - \big(\max_{j\ne j^\star} s_j + \varepsilon\big)
\;=\; \gamma - 2\varepsilon,
\]
where $\varepsilon := C\sqrt{\log(M/\delta)/(p\,s_{\mathrm{eff}}(q))}$.
Thus, if $\gamma > 2\varepsilon$, the post-dropout margin is positive and
$\arg\max_j s'_j = j^\star$, i.e., the top-1 prediction is unchanged, with probability at least $1-\delta$.

\end{document}